\documentclass[10pt]{svproc}
\usepackage{url}

\newcommand\blfootnote[1]{%
  \begingroup
    \renewcommand{\thefootnote}{} 
    \footnotetext{#1}
    \renewcommand{\thefootnote}{\arabic{footnote}}
  \endgroup
}

\usepackage{amsmath}
\usepackage{mathtools}
\usepackage{setspace}
\usepackage{lpform}
\usepackage{textcomp}
\usepackage{algorithmic}
\usepackage[english]{babel}
\usepackage{amssymb}
\usepackage{graphicx}
\usepackage{xcolor}
\usepackage{enumerate}
\usepackage{framed}
\usepackage{cite}
\usepackage{tabularx}
\usepackage{mathrsfs}
\usepackage[innerleftmargin=5pt,innerrightmargin=5pt,skipabove=5pt]{mdframed}

\let\emptyset\varnothing

\usepackage{caption}
\newcommand{\gobble}[1]{}
\newcommand{\gobblexor}[2]{#2} 

\newcommand{\secref}[1]{Section~\ref{#1}} 

\usepackage{verbatim}

\newcommand*{\problemStInternal}[4]{
	\begin{mdframed}\parbox{0.98\columnwidth}{
		 \begin{small}\textbf{#4: #1} \\[0.05in]\end{small}
			\renewcommand{\tabcolsep}{0.5pt}
			\begin{tabularx}{\linewidth}{rX}
				\emph{Input:\,} & #2 \\
				\emph{Output:\,} & #3
			\end{tabularx}
		}\end{mdframed}
		\par
	}
	
\newcommand*{\problemSt}[3]{\problemStInternal{#1}{#2}{#3}{Problem}}

\makeatletter
\newcommand*{\Relbarfill@}{\arrowfill@\Relbar\Relbar\Relbar}
\newcommand*{\xeq}[2][]{\ext@arrow 0055\Relbarfill@{#1}{#2}}
\makeatother

	 \providecolor{added}{rgb}{0,1,0}
	 \providecolor{changed}{rgb}{0,0,0}
	 \providecolor{deleted}{rgb}{1,0,0}
	 
	 \newcommand{\changed}[1]{{\color{changed}{}#1}}

\newcommand{\Lang}{\ensuremath{\mathcal{L}}} 

\usepackage[norelsize, linesnumbered, ruled, lined, boxed, commentsnumbered]{algorithm2e}

\DeclareMathOperator*{\argmin}{arg\,min} 
\newcommand\append{}

\newcommand{\RTM}{{\sc Rtm}\xspace}
\newcommand{\RTMFOM}{{\sc Rtm/Fom}\xspace}
\newcommand{\RTMFHM}{{\sc Rtm/Fhm}\xspace}

\SetKwRepeat{Do}{do}{while}%

\begin{document}

\mainmatter              
%
\title{\LARGE \bf Planning to Chronicle}
\titlerunning{Planning to Chronicle}  
%
\author{Hazhar Rahmani\inst{1} \and Dylan A. Shell\inst{2} \and Jason M. O'Kane\inst{1}}
\authorrunning{Rahmani, Shell, and O'Kane} 
%
\tocauthor{Hazhar Rahmani, Dylan A. Shell, and Jason M. O'Kane}
\institute{University of South Carolina, Department of Computer Science and Engineering\\
\and
Texas A\&M University, Department of Computer Science and Engineering
}

\maketitle              

    %
    %

\begin{abstract}
An important class of applications entails a robot monitoring, scrutinizing, or
recording the evolution of an uncertain time-extended process. This sort of
situation leads to an interesting family of planning problems in which the robot
is limited in what it sees and must, thus, choose what to pay attention to.
The distinguishing characteristic of this setting is that the robot has
influence over what it captures via its sensors, but exercises no causal
authority over the evolving process.  As such, the robot's objective is to
observe the underlying process and to produce a `chronicle' of occurrent
events, subject to a goal specification of the sorts of event sequences that
may be of interest. This paper examines variants of such problems when the
robot aims to collect sets of observations to meet a rich specification of
their sequential structure.  We study this class of problems by modeling a
stochastic process via a variant of a hidden Markov model, and specify the
event sequences of interest as a regular language, developing a vocabulary of
`mutators' that enable sophisticated requirements to be expressed.  Under
different suppositions about the information gleaned about the event model, we
formulate and solve different planning problems. The core underlying idea is
the construction of a product between the event model and a specification
automaton.  The paper reports and compares performance metrics by drawing on
some small case studies analyzed in depth in simulation. 
\keywords{Planning, Story-telling, Reconnoitering, Raconteuring}
\end{abstract}

\blfootnote{This work was graciously supported, in part, by the National Science Foundation through awards
IIS-1453652, IIS-1849249, and IIS-1849291.}

\vspace*{-6pt}  
\section{Motivation and Introduction}\label{sec:intro}
\vspace*{-8pt}
This paper is about robotic planning problems in which the goals are expressed
as time-extended sequences of discrete events whose occurrence the robot cannot
causally influence.  \changed{As a concrete motivation for this sort of setting, consider
the proliferation of home videos.}  These videos are, with remarkably few
exceptions, crummy specimens of the cinematic arts.  They fail, generally, to
establish and then bracket a scene; they often founder in emphasizing the
importance of key subjects within the developing action, and are usually
unsuccessful in attempts to trace an evolving narrative arc. And the current
generation of autonomous personal robots and video drones, in their roles as
costly and glorified `selfie sticks,' are set to follow suit. The trouble
is that capturing footage to tell a story is challenging.  \changed{A camera can only
record what you point it toward, so part of the difficulty stems from the fact
that you can't know exactly how the scene will unfold before it actually does.}
Moreover, what constitutes structure isn't easily summed up with
a few trite quantities.  Another part of the challenge, of course, is that one
has only limited time to capture video footage.

Setting aside pure vanity as a motivator, many applications can be cast as the
problem of producing a finite-length sensor-based recording of the evolution of
some process.  As the video example emphasizes, one might be interested in
recordings that meet rich specifications of the event sequences that are of
interest.  When the evolution of the event-generating process is
uncertain/non-deterministic and sensing is local (necessitating its active
direction), then one encounters an instance from this class of problem.
The broad class encompasses many monitoring and surveillance scenarios.  An important
characteristic of such settings is that the robot has influence over what it
captures via its sensors, but cannot control the process of interest.

Our incursion into this class of problem involves two lines of attack. The
first is  a wide-embracing formulation in which we pose a general stochastic model,
including aspects of hidden/latent state, simultaneity of event occurrence, and
various assumptions on the form of observability. Secondly, we specify the
sequences of interest via a deterministic finite automaton (DFA), 
and we define several language mutators, which
permit composition and refinement of specification DFAs, allowing for rich
descriptions of desirable event sequences.
The two parts are brought together via our approach to planning: we
show how to compute an optimal policy (to satisfy the specifications as quickly
as possible) via a form of product automaton. Empirical evidence from
simulation experiments attests to the feasibility of this approach.

Beyond the pragmatics of planning, a theoretical contribution of the paper is
to prove a result on representation independence of the specifications.  That
is, though multiple distinct DFAs may express the same regular language and
despite the DFA being involved directly in constructing the product automaton
used to solve the planning problem, we show that it is merely the language
expressed that affects the resulting optimal solution.  Returning to mutators
that transform DFAs, enabling easy expression of sophisticated requirements, we
distinguish when mutators preserve representational independence too. 

\vspace*{-6pt}
\section{Related Work} \label{sec:related}
\vspace*{-8pt}
Our interest in understanding robot behavior in terms of the robots'
observations of a sequence of discrete events is, of course, not unique.
The \emph{story validation problem}~\cite{yu2010cyber,yu2011story} can be
viewed as an inverse of our problem.  The aim there is to
determine whether a given story is consistent with a sequence of events
captured by a network of sensors in the environment. In our problem, it is the
robot that needs to capture a sequence of events that constitute a desired
story.

\emph{Video summarization} is the problem of making a `good' summary of a given
video by prioritizing sequences of frames based on some
selection criterion (importance, representativeness, diversity, etc.).
Various approaches include identifying important objects~\cite{lee2012discovering}, finding interesting events~\cite{gygli2014creating}, selection using supervised learning~\cite{gong2014diverse}, and finding inter-frame connections~\cite{lu2013story}. 
\changed{For a survey on video summarization see\gobble{ Truong and Venkatesh}~\cite{truong2007video}, which one might augment with the more recent results of~\cite{mahasseni2017unsupervised, plummer2017enhancing, zhang2018retrospective, ji2019video}. 
}
Girdhar and Dudek~\cite{girdhar2012efficient} considered the related \emph{vacation snapshot problem}, in which the goal is to retain a diverse subset from data observed by a mobile robot.
However, in such summarization techniques, the problem is essentially to
post-process a collection of images already recorded.  This paper, by contrast, addresses the problem of deciding which
video segments the robot should attempt to capture in the first place.

For text-based and interactive narratives, a variety of methods are known for narrative planning and generating natural language stories~\cite{riedl2010narrative, robertson2017narrative}.

\changed{Closely related research to the present paper is\gobblexor{ the work of
Shell, Huang, Becker, and O'Kane~\cite{shell2019planning}\,---having some
authors in common with the present paper---\,}{~\cite{shell2019planning}, }which introduces the idea of using
a team of autonomous robots, coordinated by a planner, to capture a sequence of
events specifying a given narrative structure. }
That work raised (but did not answer) several questions, among which is how the robot can formulate effective plans to capture events relevant to the story specification.
Here we
build upon that prior effort showing how such plans can be formed in a principled way.

Related to our problem are also the theories of Markov decision processes (MDPs) and partially observable Markov decision processes (POMDPs), which are surveyed \gobblexor{by LaValle}{in}~\cite{lavalle2006planning, shani2013survey, ross2008online, bonet2009solving}. 
\changed{We solve our problem by constructing a product of the event model and the specification, which together yield a specific POMDP.}
%
%

\vspace*{-6pt}
\section{The Problem}\label{sec:problem}
\vspace*{-8pt}
First, we introduce the basic elements of our model and problem formalization.
\vspace*{-8pt}
\subsection{Events and observations}
The essential objects of interest are \emph{events}, that is, atomic
occurrences situated at specific times and places. We propose to treat each
event as a letter drawn from a finite alphabet $E$, a set which contains all
possible events. 
Accordingly, any finite sequence of events, in particular a \emph{story} $\xi$
the robot wants to record from the events that occur in the system, is a finite
word in $E^*$.

We model the occurrence of events using a structure defined as follows.
%

\vspace*{-4pt}
\begin{definition}[event model]
\label{def:markovChain}
An \emph{event model} $\mathcal{M}=(S, \mathbf{P}, s_0, E, g)$ is a tuple in which
    (1) $S$, which is a nonempty finite set, is the \emph{state space} of the model;
    (2) $\mathbf{P}: S \times S \to [0, 1]$ is the \emph{transition
	    probability function} of the model, such that for each state $s \in
		S$, $\sum_{s' \in S}\mathbf{P}(s, s')=1$;
	(3) $s_0 \in S$ is the \emph{initial state};
    (4) $E$ is the set of all possible events; and
    (5) $g: S \to 2^E$ is a labeling function assigning, to each state, the (possibly empty) set of events that are occurring (mnemonically, `{g}oing-on') simultaneously at that state. We assume that $g(s_0) = \emptyset$.
\end{definition}
\vspace*{-6pt}
\noindent An execution of the model starts from the initial state $s_0$ and then, at each
time step $k$, the system makes a transition from state $s_k$ to state
$s_{k+1}$, the latter being chosen randomly based on $\mathbf{P}$ from those
states for which $\mathbf{P}(s_k, \cdot) > 0$.
This execution specifies a path $s_0 s_1 \cdots$.
For every time step $k$, when the system enters state $s_k$, each event in
$g(s_k)$ occurs simultaneously.

We are interested in scenarios in which a robot is tasked
with recording certain sequences of events.  We model the
state of the event model as only partially observable to the
robot.  That is, the current state $s_k$ of the event model
is hidden from the robot, but the system instead emits an
output observable to the robot at each time step.  The next
definition formalizes the idea.
%
\vspace*{-2pt}
\begin{definition}[observation model]
For a given event model $\mathcal{M}=(S, \mathbf{P}, s_0,$ $E, g)$, an \emph{observation model} $\mathcal{B}=(Y, h)$
is a pair in which
     (1) $Y$ is a set of \emph{observations} or outputs; 
     (2) $h: S \times Y \to [0, 1]$ is the \emph{emission probability function} of the model, such that for each state $s \in S$, $\sum_{y \in Y}h(s, y)=1$.
\end{definition}
\vspace*{-2pt}

At each time step, when the system enters a state $s_k$, it emits an output $y_k$, drawn according to $h(s_k,\cdot)$.
The emitted output $y_k$ is observable to the robot.
\changed{An event model and observation model can be depicted together as a directed graph (e.g., see Figure~\ref{fig:product}a), 
where we show each state's events as an attached set (in braces in the figure) and display observations from $Y$ along with their emission probabilities (in brackets).} 
We consider, as important special cases, two particular types of observation models.
\vspace*{-2pt}
\begin{definition}
  Given an event model $\mathcal{M}=(S, \mathbf{P}, s_0, E,
  g)$ with observation model $\mathcal{B}=(Y, h)$, we say that $\mathcal{B}$ makes $\mathcal{M}$ 
   \emph{fully observable} if (1) $Y=S$, and (2)~$h(s, y)=1$ if and only if $s = y$.
\end{definition}
\vspace*{-6pt}
\changed{We write $\mathcal{B}_{obs}(\mathcal{M})$ to denote the unique observation model that makes $\mathcal{M}$ fully observable.}
At the other extreme, another special event model is one in
which the emitted outputs do not help at all to reduce
uncertainty.
\vspace*{-2pt}
\begin{definition}
Given an event model $\mathcal{M}=(S, \mathbf{P}, s_0, E, g)$ with observation model $\mathcal{B}=(Y, h)$, then $\mathcal{B}$ causes the event model to be \emph{fully hidden} if the observation space $Y$ is a \changed{singleton set}.
\end{definition}
\vspace*{-4pt}
\changed{
Since the particular single observation comprising $Y$ is unimportant,  by $\mathcal{B}_{hid}(\mathcal{M})$ we denote 
some observation model making $\mathcal{M}$ fully hidden. 
}
\vspace*{-8pt}
\subsection{Story specifications, belief states, and policies}
As the system evolves along $s_0 s_1 s_2 \cdots$, the robot attempts to record some of the events that actually occur in the world to form a story $\xi \in E^*$.
We specify the desired story using a deterministic
finite automaton (DFA) $\mathcal{D}=(Q, E, \delta, q_0, F)$,
where $Q$ is its state space, $E$ is its alphabet, $\delta: Q
\times E \rightarrow Q$ is its transition function, $q_0$ its
initial state, and $F \subseteq Q$ is the set of all final
(accepting) states of the automaton.
In other words, we want the robot to make a story $\xi$ in the language of $\mathcal{D}$, denoted $\Lang(\mathcal{D})$, which is the set of all strings in $E^*$ that when are tracked from $q_0$, the automaton reaches an accepting state.

The semantics of event capture are as follows.  At each step $k \geq 0$, the robot chooses one event $e$ from $E$ to attempt to record in the next step, $k+1$. 
If any of the actual events that do happen at step $k+1$ (i.e., any of the events in $g(s_{k+1})$) match the robot's prediction, then the robot successfully records this event; otherwise, it records nothing.
The robot is aware of the success or failure of each of its attempts.
The robot stops making guesses and observations once it has recorded a desired story---a story in $\Lang(\mathcal{D})$.

To estimate the current state, the robot maintains,
at each time step $k$, a belief state $b_k: S
\rightarrow [0, 1]$, in which $\sum_{s \in S}
b_k(s)= 1$.  \changed{For each $s \in S$, $b_k(s)$
represents the probability that the event model is in
state $s$ at time step $k$, according to the
information available to the robot, including both the observations emitted directly by the event model, along with the sequence of successes or failures in recording events.}
\changed{It also maintains, for each time $k$, 
the sequence $\xi_k$ of events it has recorded until time step $k$, and
the (unique) DFA state $q_k$ obtained by $\xi_k$.}

The robot's predictions are governed by a policy $\pi: \Delta(S) \times Q \to E$ that depends on the belief state and the state of the DFA.
At time step $k+1$, the robot may append a recorded event to $\xi_k$ via the following formula:

\vspace*{-6pt}
{\small
            \begin{equation}
            \xi_{k+1} =
			\begin{cases}
			\xi_k \append \pi(b_k, q_k) & \pi(b_k, q_k) \in g(s_{k+1})  \\
			
			\xi_k & \pi(b_k, q_k) \notin g(s_{k+1}).
			
			\end{cases}    
            \end{equation}
}
\vspace*{-8pt}            

The initial condition is that $\xi_0 = \epsilon$, in which $\epsilon$ is the empty string.
The robot changes the value of variable $q_k$ only when the guessed event actually happened: 

\vspace*{-6pt}
{\small
            \begin{equation}
            q_{k+1} =
			\begin{cases}
			\delta(q_k, \pi(b_k, q_k)) & \pi(b_k, q_k) \in g(s_{k+1}) \\
			
			q_k & \pi(b_k, q_k) \notin g(s_{k+1}).
            
			\end{cases}    
            \end{equation}
}
\vspace*{-8pt}                    

The robot stops when $q_k \in F$.
\vspace*{-8pt}
\subsection{Optimal recording problems}
The robot's goal is to record a story (or video) as quickly
as possible.  We consider this problem in three
different settings: a general setting without any
restriction on the event model, a setting in which
the event model is fully observable, and a final one
in which the event model is fully hidden.
First, the general setting.

\problemSt{Recording Time Minimization (\RTM)}
  {An event set $E$,
  an event model $\mathcal{M}=(S, \mathbf{P}, s_0, E, g)$
  with observation model $\mathcal{B}=(Y, h)$,
  and a DFA $\mathcal{D}=(Q, E, \delta, q_0, F)$.}
  {A policy minimizing the expected number of steps $k$ until $\xi_k \in \Lang(\mathcal{D})$.}

Note that $k$ is not necessarily the length of
the resulting story $\xi_k$, but rather is the number of
steps the system runs to capture that story.  In fact, $|\xi_k| \leq k$.

The second setting constrains the system to be
fully observable.

\problemSt{RTM with Fully Observable Model (\RTMFOM)}
  {An event set $E$,
  an event model $\mathcal{M}=(S, \mathbf{P}, s_0, E, g)$,
  and a DFA $\mathcal{D}=(Q, E, \delta, q_0, F)$.}
  {A policy that, under observation model
  $\mathcal{B}_{obs}(\mathcal{M})$, minimizes the
  expected number of steps $k$ until $\xi_k \in
  \Lang(\mathcal{D})$.}
  
\changed{In this setting, because states are fully observable to
the robot, we might have defined
the policy as a function over $S \times Q$ rather
than over $\Delta(S) \times Q$.  Nonetheless, our
current definition does not pose any problem.
Any reachable belief state in this setting
considers only a \emph{single} outcome (i.e., given any $k$,
$b_{k}(s)=1$ for exactly one $s \in S
$) and thus, we are interested in
the optimal policy only for those reachable beliefs.}

The third setting assumes a fully hidden event model state.
\problemSt{RTM with Fully Hidden Model (\RTMFHM)}
  {An event set $E$,
  an event model $\mathcal{M}=(S, \mathbf{P}, s_0, E, g)$,
  and a DFA $\mathcal{D}=(Q, E, \delta, q_0, F)$.}
  {A policy that, under observation model
  $\mathcal{B}_{hid}(\mathcal{M})$, minimizes the
  expected number of steps $k$ until $\xi_k \in
  \Lang(\mathcal{D})$.}

\vspace*{-6pt}
\section{Algorithm Description} \label{sec:alg}
\vspace*{-8pt}
Next we give an algorithm for \RTM, which also solves \RTMFOM and \RTMFHM, which are basically the same \RTM problem but with two special kinds of event models as inputs to the problem.
\vspace*{-8pt}
\subsection{The Goal POMDP}
\vspace*{-6pt}
The first step of the algorithm constructs a specific partially observable Markov decision process (POMDP), which we term the Goal POMDP, as follows:
\vspace*{-2pt}
\begin{definition}[Goal POMDP]
\label{def:prod}
For an event model $\mathcal{M}=(S, \mathbf{P}, s_0, E, g)$ with observation model $\mathcal{B}=(Y, h)$,
and a DFA $\mathcal{D}=(Q, E, \delta, q_0, F)$, 
the associated \emph{Goal POMDP} is a tuple $\mathcal{P}_{(\mathcal{M},\mathcal{B};\mathcal{D})}=(X, A, b_0, \mathbf{T},$ $X_G, Z, \mathbf{O}, c)$, in which 

\vspace*{-6pt}
\begin{enumerate}
    \item $X = S \times Q$ is the state space;
    \item $A = E$ is the action space;
    \item $b_0 \in \Delta(X)$ is the initial belief state, in which $b_0(x) = 1$ iff $x = (s_0, q_0)$;
    \item \label{itm:T} $\mathbf{T}: X \times A \times X \rightarrow [0, 1]$ is the transition probability function such that for each $e \in E$ and $(s, q), (s', q') \in X$,
    \vspace*{-10pt}
     $$\mathbf{T}((s, q), e, (s', q')) =
     {\small
    \begin{cases}
        \mathbf{P}(s, s') & \mbox{\text{if} $q \notin F, q' = \delta(q, e), \text{and } e \in g(s')$} \qquad\hfill (\ref{itm:T}.a) \\
        \mathbf{P}(s, s') & \mbox{\text{if} $q \notin F, q' = q, \text{and } e \notin g(s')$} \qquad\hfill (\ref{itm:T}.b) \\
        1 & \mbox{if $q \in F, q' = q, \text{and } s = s'$} \qquad\hfill (\ref{itm:T}.c) \\
        0 & \text{otherwise};
    \end{cases}
    }
     $$
     \vspace*{-16pt}
     \item $X_G = S \times F$ is the set of goal states;
     \item $Z = \left ( \lbrace \mathrm{True}, \mathrm{False} \rbrace \times Y \right ) \cup \lbrace {\bot} \rbrace$ is the set of observations;
     \item \label{itm:O} $\mathbf{O}: A \times X \times Z \rightarrow [0, 1]$ is the observation probability function such that for each $e \in E$, $s \in S$, $q \in Q$, and $y \in Y$:
     \begin{enumerate}[{(}a{)}]
         \item \label{itm:O_1} $\mathbf{O}(e, (s, q), (\mathrm{True}, y)) = h(s, y)$ if $q \notin F$ and $e \in g(s)$,
         \item \label{itm:O_1p} $\mathbf{O}(e, (s, q), (\mathrm{False}, y)) = 0$ if $q \notin F$ and $e \in g(s)$,
         \item \label{itm:O_2} $\mathbf{O}(e, (s, q), (\mathrm{False}, y)) = h(s, y)$ if $q \notin F$ and $e \notin g(s)$,
         \item \label{itm:O_2p} $\mathbf{O}(e, (s, q), (\mathrm{True}, y)) = 0$ if $q \notin F$ and $e \notin g(s)$,
         \item \label{itm:O_3} $\mathbf{O}(e, (s, q), \bot) = 1$ if $q \in F$;
     \end{enumerate}

     \item $c: X \times A \rightarrow \mathbb{R}_{\geq 0}$ is the cost function such that for each $x \in X$ and $a \in A$, $c(x, a)=1$ if $x \notin X_G$, and $c(x, a)=0$ otherwise.
\end{enumerate}
\end{definition}

\noindent Each state of this POMDP is a pair $(s, q)$ indicating the situation where, under an execution of the system, the current state of the event model is~$s$ and the current state of the DFA is~$q$. 
For each $x, x' \in X$ and $a \in A$, $\mathbf{T}(x, a, x')$ gives the probability of transitioning from state $x$ to state $x'$ under performance of action~$a$.
In the context of our event model, each transition corresponds to a situation where the robot chooses an event $e$ to observe and the event model makes a transition from a state $s$ to~$s'$. 
If $e$ appears in $\mathrm{g}(s')$, then the robot records $e$ and then changes the current state of the DFA to $\delta(q, e)$; otherwise, it does nothing and the DFA remains in state~$q$.
These correspond to cases (\ref{itm:T}.a) and (\ref{itm:T}.b) above, respectively. Case~(\ref{itm:T}.c) makes all the goal states of the POMDP \emph{absorbing} states.
The goal states of the POMDP are those in which the robot has recorded a story, i.e., the current state of the specification DFA is accepting.

For each $a \in A$, $x \in X$, and $z \in Z$, the function $\mathbf{O}(a, x, z)$ is an observation model, its value being the probability of observing $z$ given that the system has entered state $x$ via action $a$.
The POMDP has a special observation, $\bot$, which is observed only when a goal state is reached.
Any other observation is a pair $(r, y)$ where $r \in \lbrace \mathrm{True}, \mathrm{False} \rbrace$ discloses whether the robot's prediction was correct---the event did happen---or not, and $y$ indicates the sensed observation the robot made (as per $\mathcal{B}$).
Rules~\ref{itm:O_1}--\ref{itm:O_2p} ensure that the first element of the observation pair informs the robot whether its prediction was correct.
To see this, if the robot has predicted $e$ to occur, the event model has entered state $s$ such that $e \in g(s)$, and the robot has made an observation $y$, then the probability of observing $(\mathrm{True}, y)$ by entering to state $(s, q)$ via action $e$ is equal $h(s, y)$ (case~\ref{itm:O_1}). 
If event $e \notin g(s)$, then the robot's prediction has to be wrong, and thus, the probability of observing $(\mathrm{False}, y)$ in state $(s, q)$ when it is reached via action $e$ is $h(s, y)$ (expressed in case~\ref{itm:O_2}). Cases~\ref{itm:O_1p} and~\ref{itm:O_2p} ensure that there is no misreporting of the correctness of the prediction.
Case~\ref{itm:O_3} indicates the observation that the robot has completed recording of a story in $\mathcal{L}(\mathcal{D})$.

\begin{figure}[t]
  \centering
  \includegraphics[width=\linewidth]{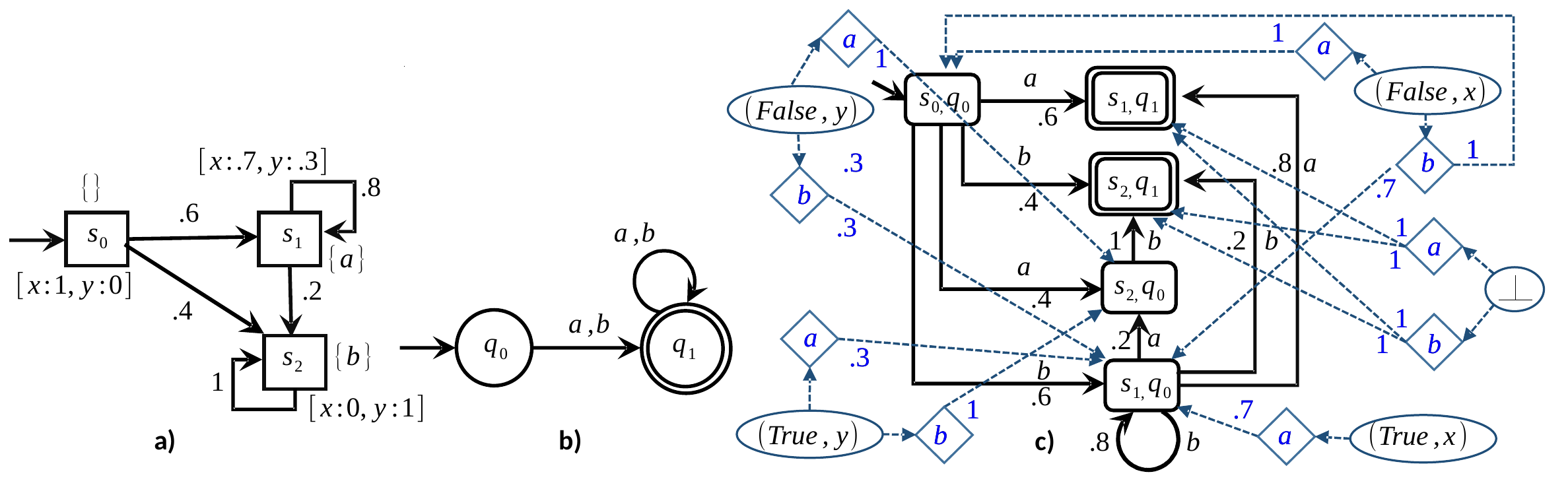}
  \caption{
    \textbf{a)} An event model $\mathcal{M}$ with its observation model $\mathcal{B}$.
    \textbf{b)} A DFA $\mathcal{D}$, specifying event sequences that contain at least one event.
    \textbf{c)} The Goal POMDP 
    $\mathcal{P}_{(\mathcal{M},\mathcal{B};\mathcal{D})}$, constructed by Definition~\ref{def:prod}. (Self-loop transitions of the goal states have been omitted to try reduce visual clutter.)
    }
    \vspace*{6pt}
  \label{fig:product}
\end{figure}

Figure~\ref{fig:product} illustrates this construction for an elementary example.

\vspace*{-8pt}
\subsection{Solving the Goal POMDP}
\changed{
A POMDP is usually formulated as a fully observable MDP called a \emph{belief MDP} whose (continuous) state space consists of the belief space of the POMDP. 
Accordingly, in the belief MDP from Goal POMDP $\mathcal{P}=(X, A, b_0, \mathbf{T},$ $X_G, Z, \mathbf{O}, c)$, for each belief state $b \in \Delta(X)$, action $a \in A$, and observation $z \in Z$,  
we denote the updated belief state of $b$ after action $a$ and observation $z$ by $b_z^a$. It is computed as follows:}
\vspace*{-8pt}
\begin{equation}
\label{eq:b_za}
b_{z}^a(x) = Pr(x|z, a, b) = \frac{\mathbf{O}(a, x, z)\sum_{x' \in X}\mathbf{T}(x', a, x)b(x')}{Pr(z| a, b)},    
\end{equation}
in which, 
\vspace*{-8pt}
\begin{equation}
\label{eq:Pr_z}
  Pr(z| a, b) = \sum_{x \in X}\mathbf{O}(a, x, z) \sum_{x' \in X}\mathbf{T}(x', a, x)b(x').
\end{equation}
For this belief MDP, the cost of each action $a$ at belief state $b$ is $c'(b, a) = \sum_{x \in X}b(x)c(x, a)$, which in our case, $c'(b, a) = 1$ if $b$ is a not a goal belief state, and otherwise $c'(b, a) = 0$.
An optimal policy ${\pi'}^*: X \rightarrow A$ for this MDP is formulated as a solution to the Bellman recurrences
{\small
\vspace*{-4pt}
\begin{equation}
\label{eq:GStar}
    {V'}^*(b) = \min\limits_{a \in A} \big(c'(b, a)+ \sum\limits_{z \in Z} Pr(z | a, b) {V'}^*(b_z^a)\big),
\end{equation}
\vspace*{-8pt}
\begin{equation}
\label{eq:PiStar}
    {\pi'}^*(b) = \argmin\limits_{a \in A} \big(c'(b, a)+ \sum\limits_{z \in Z} Pr(z | a, b){V'}^*(b_z^a) \big).
\end{equation}}%
One can use any standard technique to solve these recurrences. For surveys on methods, see~\cite{bonet2009solving, shani2013survey, ross2008online}.
An optimal policy computed via these recurrences prescribes, for any belief state reachable from $b_0$, an optimal action to execute.
Accordingly, the robot executes at each step, the action given by the optimal policy, and then updates its belief state via \eqref{eq:b_za}.
One can show, via induction, that at each step $i$, there is a unique $q_i \in Q$ such that belief state $b_i$ has outcomes only for (but probably not all) $x_j=(s_j, q_i) \in X, j = 1, 2,  \cdots |S|$.
As such, function $\beta: \Delta(X) \rightarrow \Delta(S) \times Q$ maps each $b_i$ of those belief states to a tuple $(d, q_i)$, where for each $s \in S$, $d(s)=b((s, q_i))$.
Subsequently, the optimal policy ${\pi'}^*$ computed for $\mathcal{P}_{(\mathcal{M}, \mathcal{B}; \mathcal{D})}$ can be mapped to an optimal solution $\pi^*: \Delta(S) \times Q \rightarrow A$ to \RTM, 
by interpreting $\pi^*(\beta(b_i))={\pi'}^*(b_i)$, for each reachable belief state $b_i \in \Delta(X)$. 

 \vspace*{-8pt}
\subsection{Solving \RTMFOM via Goal MDP}
\label{sec:goalMDP}
\vspace*{-2pt}
The previous construction can be used to solve \RTMFOM too, but, given that the event model fed into \RTMFOM is fully observable, to improve solution tractability, it is more sensible to construct a Goal MDP.
To do so, for the event model and the DFA in Definition~\ref{def:prod}, the Goal MDP $\mathcal{M}=(X, A, b_0, \mathbf{T},$ $X_G, c)$ embedded in the POMDP $\mathcal{P}$ in that definition is extracted and 
then an optimal policy for $\mathcal{M}$ is solved.
An optimal policy ${\pi''}^*$ for the MDP is a function over $X=S \times Q$, which is computed via the Bellman equations

\vspace*{-4pt}
{\small
\begin{equation}
\label{eq:GStar}
    {V''}^*(x) = \min\limits_{a \in A} \big(c(x, a)+ \sum\limits_{x'\in X} {V''}^*(x') \mathbf{T} (x, a, x') \big),
\end{equation}
\vspace*{-6pt}
\begin{equation}
\label{eq:PiStar}
    {\pi''}^*(x) = \argmin\limits_{a \in A} \big(c(x, a)+ \sum\limits_{x'\in X} {V''}^*(x') \mathbf{T}(x, a, x')\big).
\end{equation}%
\vspace*{-2pt}%
}%
\changed{%
These equations may be solved by a variety of methods (see \cite[Chp.~10]{lavalle2006planning} for a survey).
In the evaluation reported below, we use standard value iteration.
After computing $\pi{''}^*$, for each $x = (s, q) \in X$, we make a belief state $b \in \Delta(S)$ such that $b(s') = 1$ if and only if $s' = s$, and then set $\pi^*(b, q) = {\pi''}^*((s, q))$, where $\pi^*$ is an optimal solution to \RTMFOM.
Observe that $\pi^*$ for \RTMFOM is only computed for finitely many pairs $(b, q)$, those in which $b$ is a single outcome.
}

\vspace*{-6pt}
\section{Representation-invariance of expected time} \label{sec:repInv}
\vspace*{-8pt}
Notice that the event selected by the policy $\pi^*$ at each step depends, in part, on the current state of the specification DFA.
Because a single regular language may be represented with a variety of
distinct DFAs with different sets of states ---and thus, their optimal policies
cannot be identical--- one might wonder whether the expected execution
time achieved by their computed policies depends on the specific DFA,
rather than only on the language.
The question is particularly relevant in light of the language mutators we examine in~\secref{sec:mutator}.
Here, we show that the expected number of steps required to capture a story
within a given event model does indeed depend only on the language specified
by the DFA, and not on the particular representation of that language.

For a DFA $\mathcal{D}=(Q, E, \delta, q_0, F)$, we define a function $f: Q \rightarrow \lbrace 0, 1 \rbrace$ such that for each $q \in Q$, $f(q) = 1$ if $q \in F$, and otherwise, $f(q) = 0$.
Now consider the well-known notion of bisimulation, defined as follows:
\vspace*{-2pt}
\begin{definition}[bisimulation\cite{rot2016proving}]%
\label{def:bisim}%
Given DFAs $\mathcal{D}=(Q, E, \delta, q_0, F)$ and $\mathcal{D}'=(Q', E, \delta', q_0', F')$, a relation $R \subseteq Q \times Q'$ is a \emph{bisimulation relation} for $(\mathcal{D}, \mathcal{D}')$ if for any $(q, q') \in R$:
    (1) $f(q) = f'(q')$;
    (2) for any $e \in E$, $(\delta(q, e), \delta'(q', e)) \in R$.
\end{definition}
\vspace*{-4pt}
Bisimulation implies language equivalence and vice versa.
\vspace*{-2pt}
\begin{proposition}%
\label{prop:simDFA}%
\emph{\hbox{(\!\!\cite{rot2016proving})}}
For two DFAs $\mathcal{D}=(Q, E, \delta, q_0, F)$, $\mathcal{D}'=(Q', E, \delta',q_0',F')$, we have $\Lang(\mathcal{D}) = \Lang(\mathcal{D}')$ iff $(q_0, q_0') \in R$ for a bisimulation relation $R$ for $(\mathcal{D}, \mathcal{D}')$.
\end{proposition}
\vspace*{-4pt}
Bisimulation is preserved for any reachable pairs. 
\changed{The state to which a DFA with transition function $\delta$ reaches by tracking an event sequence $r$ from state $q$ is denoted $\delta^*(q, s)$.}
\vspace*{-2pt}
\begin{proposition}
\label{prop:bisimReachable}
If $(q, q')$ are related by a bisimulation relation $R$ for $(\mathcal{D}, \mathcal{D}')$, then for any $r \in E^*$, $(\delta^*(q, r), {\delta'}^*(q', r)) \in R$.
\end{proposition}
\vspace*{-4pt}
We now define a notion of equivalence for a pair of belief states.
\vspace*{-2pt}
\begin{definition}
\label{def:eqBelief}
Given an event model $\mathcal{M}=(S, \mathbf{P}, s_0, E, g)$, an observation model $\mathcal{B}=(Y, h)$ for $\mathcal{M}$, DFAs $\mathcal{D}=(Q, E, \delta, q_0, F)$ and
$\mathcal{D'}=(Q', E, \delta', q_0', F')$ such that $\Lang(\mathcal{D}) = \Lang(\mathcal{D}')$, let $\mathcal{P}_{(\mathcal{M}, \mathcal{B}; \mathcal{D})}= (X, A, b_0, \mathbf{T}, X_G, Z, \mathbf{O}, c)$ and $\mathcal{P}'_{(\mathcal{M}, \mathcal{B}; \mathcal{D}')} = (X', A, b_0', \mathbf{T}', X_G', Z, \mathbf{O}', c')$. For two reachable belief states $b \in \Delta(X)$ and 
$b' \in \Delta(X')$, 
with $\beta(b) = (d, q)$ and $\beta'(b') = (d', q')$,  
we say that $b'$ 
is \emph{equivalent} to $b$,
denoted $b \equiv b'$,
if (1)
$(q, q')$ are related by a bisimulation relation for $(\mathcal{D}, \mathcal{D}')$ and that (2) $d = d'$, i.e. for each $s \in S$, $d(s) = d'(s)$.
\end{definition}
\vspace*{-4pt}
Equivalence is preserved for updated belief states.
\vspace*{-2pt}
\begin{lemma}
\label{lem:eqBelief}
Given the structures in Definition~\ref{def:eqBelief}, let $b \in \Delta(X)$ and $b' \in \Delta(X')$ be two reachable belief states such that $b \equiv b'$. For any action $a \in A$ and observation $z \in Z$, it holds that $b_z^a \equiv {b'}_z^a$ and that $Pr(z | a, b) = Pr(z | a, b').$
\end{lemma}
\vspace*{-4pt}

 Note that for a Goal POMDP $\mathcal{P}$ with initial belief state $b_0$, $V^*(b_0)$ is the expected cost of reaching a goal belief state by an optimal policy for $\mathcal{P}$. We now present our result.
\vspace*{-2pt}
\begin{theorem}
\label{thr:eq}

For the structures in Definition~\ref{def:eqBelief}, 
it holds that $V^*(b_0) = {V'}^*(b_0')$.
\end{theorem} 
\vspace*{-8pt}
\begin{proof}
\changed{
    For a belief MDP $\mathcal{M}$, let $Tree(\mathcal{M})$ to be its tree-unravelling---the tree whose paths from the root to the leaf nodes are all possible paths in $\mathcal{M}$ that start from the initial belief state.
    A policy $\pi$ for $\mathcal{M}$ chooses a fixed set of paths over $Tree(\mathcal{M})$, and the expected cost of reaching a goal belief state under $\pi$ is equal to $\sum_{p \in GoalPaths(\pi, Tree(\mathcal{M}))} C(p)*W(p)$, where $GoalPaths(\pi, Tree(\mathcal{M}))$ is the set of all paths that are chosen by $\pi$ and reach a goal belief state from the root of $Tree(\mathcal{M})$, $C(p)$ is the sum of costs of all transitions in path $p$, and $W(p)$ is the product of the probability values of all transitions in $p$.
    The idea is that if we can overlap the tree-unravellings of the belief MDPs $\mathcal{P}_{(\mathcal{M}, \mathcal{B}; \mathcal{D})}$ and $\mathcal{P}'_{(\mathcal{M}, \mathcal{B}; \mathcal{D}')}$ in such a way that each pair of overlapped belief states are equivalent in the sense of Definition~\ref{def:eqBelief} and that each pair of overlapped transitions have the same probability and the same cost, then for each pair of overlapped belief states $b \in \Delta(X)$ and $b' \in \Delta(X')$, if we use $\pi^*(b)$ as the decision at the belief state $b'$, then because those fixed paths are overlapped, then $V^*(b_0) \geq {V'}^*(b_0')$. And, in a similar fashion, $V^*(b_0) \leq {V'}^*(b_0')$, and thus, $V^*(b_0) = {V'}^*(b_0')$.
    The following construction makes those trees and shows how we can overlap them.

    For an integer $n \geq 1$, we can make two trees $T_n$ and $T'_n$ by the following procedure.
    (1) Set $b_0$ as the root of $T_n$ and set $b_0'$ as the root of $T_n'$; make a relation $R$ and set $R \gets \lbrace (b_0, b_0') \rbrace$. 
    (2) While $|T_n| < n$, extract a pair $(b, b')$ from $R$ that has not been checked yet and in which $b$ and $b'$ are not goal belief states;
    for each action $a$ and observation $z$, compute $b_z^a$ and ${b'}_z^a$, add node $b_z^a$ and edge $(b, b_z^a)$ to $T$, and add node ${b'}_z^a$ and edge $(b', {b'}_z^a)$ to $T'$; label both edges $(a, z)$.
  Also assign to edge $(b, b_z^a)$, $Pr(z|a, b)$ as its probability value, and set the probability value of $(b^\prime, {b^\prime}_a^z)$, $Pr(z|a, b^\prime)$; the cost of each edge is set 1.
    Given that $\Lang(\mathcal{D})=\Lang(\mathcal{D}')$, by Proposition~\ref{prop:simDFA}, states $q_0$ and $q_0'$ are related by a bisimulation relation for $(\mathcal{D}, \mathcal{D}')$, which by Definition~\ref{def:eqBelief} and the construction in Definition~\ref{def:prod} implies that $b_0 \equiv b_0'$.
    This combined with Lemma~\ref{lem:eqBelief} implies that for each pair $(b, b') \in R$, $b \equiv b'$.
    We now overlap $T_n$ and $T_n'$ such that each pair $(b, b')$ that are related by $R$ are overlapped. 
    By Lemma~\ref{lem:eqBelief}, each pair of overlapped edges have the same probability value and the same cost value.
    Since for any integer $n \geq 0$ we can overlap trees $T_n$ and $T_n^\prime$ in the desired way, we can overlap the tree-unravellings of the belief MDPs of $\mathcal{P}_{(\mathcal{M}, \mathcal{B}; \mathcal{D})}$ and $\mathcal{P}_{(\mathcal{M}, \mathcal{B}; \mathcal{D}')}'$ in the desired way too; this completes the proof.

}
\end{proof}

\vspace*{-4pt}
The upshot of this analysis is that we can indeed attend only to the story
specification language (given indirectly via $\mathcal{D}$) and that the specific
presentation of that language does not impact the expected number of steps to
capture an event sequence satisfying that specification.

\vspace*{-6pt}
\section{Construction of Specification Languages}\label{sec:mutator}
\vspace*{-8pt}
In this section we describe how one
might construct, in a partially automated way, specifications for a variety of
interesting scenarios.
The idea is to use a variety of mutators to construct specification DFAs.
\newcommand{\mutator}[1]{\par\medskip\noindent{\textbf{(#1)}} }
\vspace*{-8pt}
\subsection{Multiple recipients}

Suppose we would like to capture several videos, one for each of several recipients,
within a single execution.
Given language specifications $\mathcal{D}_1,\ldots,\mathcal{D}_n \in \mathscr{D}$ \gobble{from $n$
potential recipients}, where $\mathscr{D}$ denotes the set of all DFAs over a fixed event set $E$, how can we form a single specification that directs the
robot to capture events that can be post-processed into the individual output
sequences?
One way is via two relatively simple operations on DFAs:
\vspace*{-3pt}

\mutator{$\mathbf{M_S}$} A \emph{supersequence} operation $\mathbf{M_S}: \mathscr{D} \to \mathscr{D}$, where
\vspace*{-6pt}
\begin{equation}
    \Lang(\mathbf{M_S}(\mathcal{D})) = \lbrace w \in E^* \mid \exists w' \in \Lang(\mathcal{D}), \text{$w'$ is a subsequence of $w$} \rbrace.
\end{equation}
\vspace*{-16pt}

This operation \gobblexor{might be implemented}{is produced} by first treating $\mathcal{D}$ as a
nondeterministic finite automaton (NFA), then for each event and state, adding
a transition labeled by that event from that state to itself, and \gobblexor{finally
converting the resulting NFA back into DFA form~\cite{rabin1959finite}.}{converting result back into a DFA~\cite{rabin1959finite}.}

\mutator{$\mathbf{M_I}$} An \emph{intersection} operation
$\mathbf{M_I}: \mathscr{D} \times \mathscr{D} \to \mathscr{D}$, under which 
    
        $\Lang(\mathbf{M_I}(\mathcal{D}_1, \mathcal{D}_2)) = \Lang(\mathcal{D}_1) \cap \Lang(\mathcal{D}_2)$.

\medskip\noindent
Based on these two operations, we can form a specification that asks the robot to capture an event sequence that satisfies all $n$ recipients as follows:

\vspace*{-8pt}
    \begin{equation}
        \mathcal{D} =
            \mathbf{M_I}(
                \mathbf{M_I}(\mathbf{M_S}(\mathcal{D}_1), \mathbf{M_S}(\mathcal{D}_2))
                \dots,
                \mathbf{M_S}(\mathcal{D}_n)
            )
    \end{equation}
\vspace*{-16pt}

\noindent Then from any $\xi \in \Lang(\mathcal{D})$, we can produce a $\xi_i \in \Lang(\mathcal{D}_i)$ by discarding (as a post-production step) some events from~$\xi$.

\vspace*{-8pt}
\subsection{Mistakes were made}

What should  

the robot do if it \gobble{simply }cannot capture an event
sequence that fits its specification $\mathcal{D}$, either because some
necessary events did not occur, or because the robot failed to capture them
when they did occur?
One possibility\gobble{, of course,} is to accept 
\gobble{that there will be }some limited deviation between the desired specification and \gobble{the reality of }what the robot actually captures.

Let $d: E^* \times E^* \to \mathbb{Z}^+$ denote the Levenshtein distance~\cite{levenshtein1966binary}, that is, a distance metric that measures the minimum number of insert, delete, and substitute operations needed to transform one string into another. A mutator that allows a bounded amount of such distance might be:

\vspace*{-4pt}
\mutator{$\mathbf{M_L}$} A \emph{Levenshtein mutator} $\mathbf{M_L}: \mathscr{D} \times \mathbb{Z}^+ \to \mathscr{D}$ that transforms a DFA $\mathcal{D}$ into one that accepts strings within a given distance from some string in $\Lang(\mathcal{D})$.

\vspace*{-8pt}
    \begin{equation}
        \Lang(\mathbf{M_L}(\mathcal{D}, k))
        = \{
            \xi
        \mid
            \exists \xi' \in \Lang(\mathcal{D}), d(\xi, \xi') \le k
        \}.
    \end{equation}
\vspace*{-16pt}

This mutation can be achieved using a \emph{Levenshtein automaton} construction~\cite{schulz2002fast,konstantinidis2007computing}.
Then, if the robot captures a sequence in $\Lang(\mathbf{M_L}(\mathcal{D}, k))$, it
can be converted to a sequence in $\Lang(\mathcal{D})$ by at most $k$ edits.  For
example, an insertion edit would perhaps require the undesirable use of
alternative `stock footage', rendering of appropriate footage synthetically, or
simply a leap of faith on the part of the viewer.
By assigning the costs associated with each edit appropriately in the construction, we can model the relative costs of these kinds of repairs.

\vspace*{-8pt}
\subsection{At least one good shot}
In some scenarios, there are multiple distinct views available of the same
basic event.  
We may consider, therefore, scenarios
in which this kind of good/better correspondence is known between two events,
and in which the robot should endeavor to capture, say, at least one better
shot from that class.  We define a mutator that produces such a DFA:
\vspace*{-4pt}
\mutator{$\mathbf{M_G}$} An \emph{at-least-$k$-good-shots} mutator
$\mathbf{M_G}: \mathscr{D} \times E \times E \times \mathbb{Z}^+ \to \mathscr{D}$, in
which $\mathbf{M_G}(\mathcal{D}, e, e', k)$ produces a DFA in which
$e'$ is considered to be a superior version of event $e$, and the
resulting DFA accepts strings similar to those in $\Lang(\mathcal{D})$, but with at
least $k$ occurrences of $e$ replaced with $e'$.

The construction makes a DFA in which $\mathcal{D}$ has been copied
$k+1$ times, each called a \emph{level}, with the initial state at level~$1$ and the accepting
states at level~$k+1$. Most edges remain unchanged, but each edge labeled $e$,
at all levels less than $k+1$, is augmented by a corresponding edge labeled $e'$
that moves to the next level.
\gobblexor{In this way, the specification guarantees that the superior version}{This guarantees that} $e'$ has replaced $e$ at least $k$ times, before any accepting state can be reached.

\vspace*{-6pt}
\section{Case studies}\label{sec:case}
\vspace*{-8pt}
\changed{
In this section, we present two examples solved by a Python implementation of our algorithm.
For \RTMFOM we form the Goal MDP, while for \RTMFHM and \RTM we form a Goal POMDP.
To solve the POMDP, we use APPL online (Approximate POMDP Planning Online) toolkit, which implements the DESPOT algorithm~\cite{somani2013despot}---one of the fastest known online solvers.
We compare the results for different observability conditions based upon the expected number of steps the system runs until the robot records a desired story under an optimal policy.
}
\vspace*{-8pt}
\subsection{Turisti Oulussa}
\begin{figure*}[t]
  \centering
  \includegraphics[width=\linewidth]{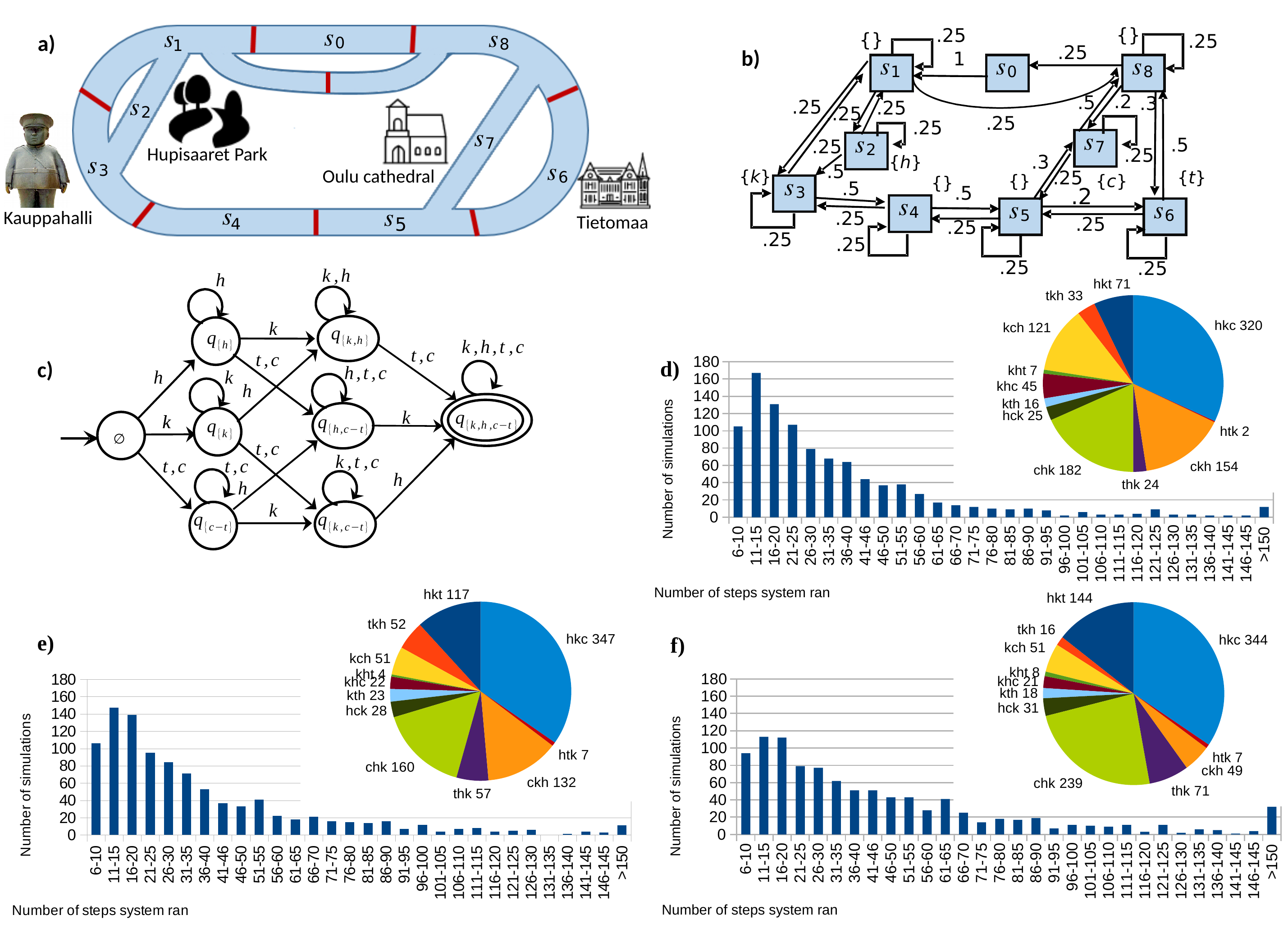}
  \caption{
  \textbf{a)} Districts of Oulu that William is touring. 
  \textbf{b)} An event model describing how a tourist visit those districts. Edges are labeled with transition probabilities.
  \textbf{c)} A DFA specifying that the captured story must contain events $k$ and $h$ and at least one of $c$~or~$t$.
    \textbf{d)} A histogram showing for a thousand simulations, the distribution of the number of hours (steps) William (system) circulated (ran) until, under the full observability assumption---the \RTMFOM problem---the robot recorded a story specified by the DFA, and a pie chart showing the distribution of recorded sequences in these simulations.
  \textbf{e)}~Histogram and pie chart for 1,000 simulations of the \RTM problem 
  where the current state of event model is observable to the robot only when William is in district $s_1$.
   \textbf{f)}~Histogram and pie chart for 1,000 simulations of the \RTMFHM problem.
   }
  \label{fig:oulu}
      \vspace*{4pt}
\end{figure*}

William is a tourist visiting Oulu as shown in Figure~\ref{fig:oulu}a.
\changed{
William's family has secretly contracted a robotic videography company to record him seeing the sights, specifically the Kauppahalli ($k$), the Hupisaaret park ($h$), and either Tietomaa museum ($t$) or the Oulu Cathedral ($c$).
}
The robot does not know William's specific plans, but it does know, through some statistics, that a typical tourist moves among those districts according to the event model in Figure~\ref{fig:oulu}b.

The desired video is specified using the DFA in Figure~\ref{fig:oulu}c.
The robot is given other tasks to do aside from recording William, and thus, \gobble{it }cannot merely follow William\gobble{ through the city}; it must form a strategy that predicts which events to try to capture.
We conducted our experiments in three settings: (1)~\RTMFOM: the robot always knows the current district in which William is located, perhaps by the help of some static sensors; (2)~\RTM: when the robot does not know at which district William is currently located but there is a single useful observation, a message sent from a security guard in district $s_1$, that informs the robot that William is in district $s_1$ whenever he is there; (3)~\RTMFHM: the robot receives no direct knowledge about William's location.
We computed the optimal policy for \RTMFOM, case (1), using the Goal MDP approach in Section~\ref{sec:goalMDP}.
According to this policy, the expected number of steps to record under a optimal policy with full observability, a story satisfying the specification, is approximately 35.24.
To verify the correctness of the algorithm, we simulated the execution of this policy 1,000 times.
In each simulation, William followed a random path through the city according to the event model in Figure~\ref{fig:oulu}b, and the robot executed the computed policy to capture an event sequence satisfying the specification. 
The average number of steps to record a satisfactory sequence for those 1,000 simulations was 35.16, quite close to the expected number of steps.
Figure~\ref{fig:oulu}d shows results of those simulations in form of a histogram and a pie chart.

For cases (2) and (3), our algorithm constructed a Goal POMDP, as described by Definition~\ref{def:prod}, and supplied it to APPL to conduct 1,000 simulations.
In case (2), \RTM with a useful observation, the average number of steps to record a desired story was 37.15, while in case (3), \RTMFHM, the average number of steps was 45.32.
Note how a single observation of whether William is in $s_1$ helped the robot to record a story considerably faster than when it did not have any state information.
Even a stream of quite limited information, if chosen aptly, can be very useful to this kind of robot.
The histograms and the pie charts for these two cases are shown in Figure~\ref{fig:oulu}e and Figure~\ref{fig:oulu}f, respectively.
Note the difference between the histograms of those three settings.

\begin{figure*}[t]
  \centering
  \includegraphics[width=\linewidth]{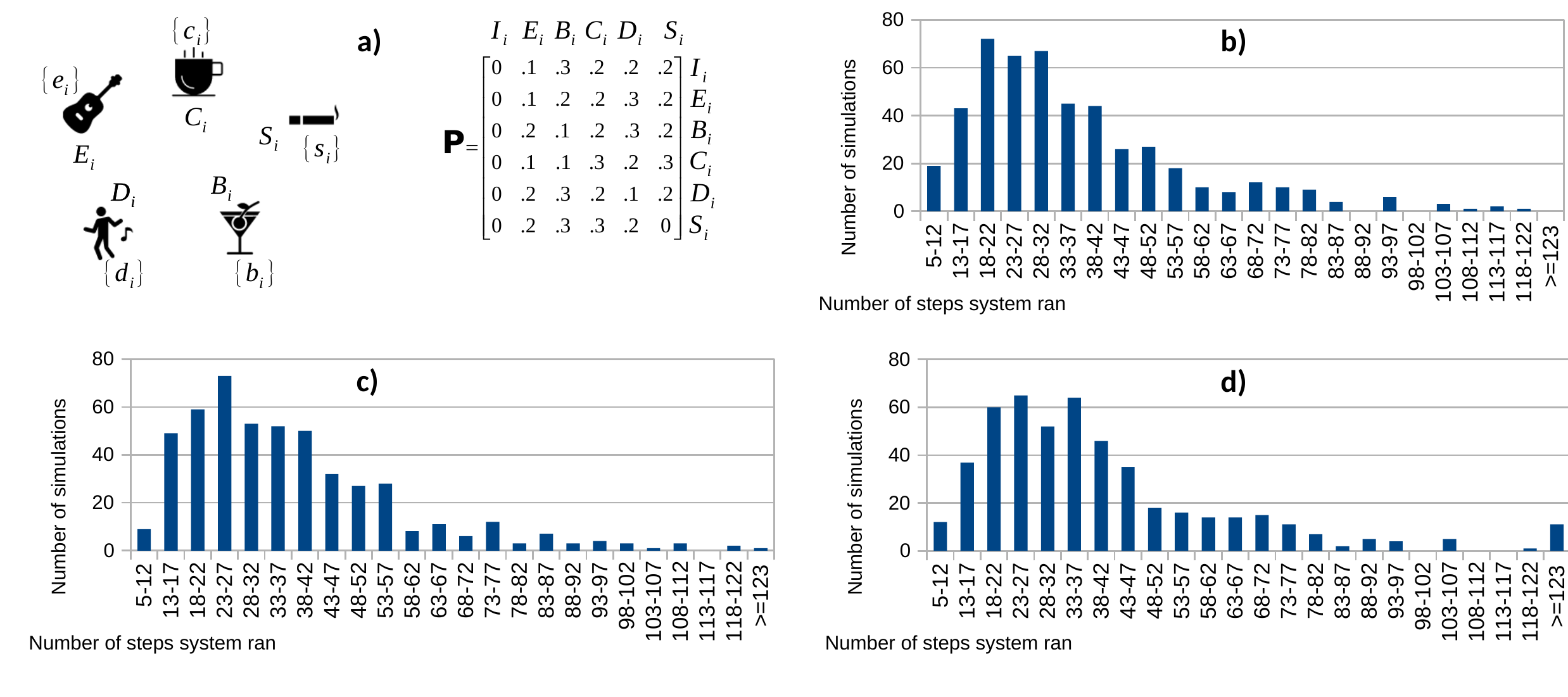}
  \caption{
  \textbf{a)} The event model for the behavior of a typical person in a party, which has six states: $I_i$, the state of arriving;
    $E_i$, the state of being entertaining;
    $C_i$, for consuming coffee;
    $B_i$, for drinking other beverages;
    $D_i$, for dancing; and
    $S_i$, for smoking.
  \textbf{b)} The histogram  of  execution  times  for 500 simulations of the wedding reception example for \RTMFOM
  \textbf{c)} The histogram  of  execution  times  for 500 simulations of the wedding reception example for \RTM with only one single observation of whether Chris is currently smoking or not
    \textbf{d)} The histogram  of  execution  times  for 500 simulations of the wedding reception example for \RTMFHM
   }
  \label{fig:party}
      \vspace*{6pt}
\end{figure*}

\vspace*{-16pt}
\subsection{Wedding reception}
A videographer robot is asked to produce videos that convey different stories, assembled from unpredictable events at a wedding reception.
  \changed{The wedding guests include Alice, Bob, and Chris, and the events of interest for any of those guests are: arriving at the reception, (i); dancing, (d); drinking coffee, (c); drinking other beverages, (b); smoking, (s); and being entertained, (e).}
  Each guest has their own sense of the events they would like to see captured: Alice is mainly interested in seeing Chris drinking or smoking, but also has plans to share the last dance with Bob; Bob cares for nothing but seeing his own dancing through the evening, but hopes to share the last dance with Alice; Chris does not care to see any events at all, but Chris's children are concerned about his unhealthy habits, and so if Chris is drinking too much coffee or smoking too much, they would like to know.
The robot in that scenario is given three parallel objectives. We can formalize those as languages, shown here for compactness as regular expressions:
for Alice, $r_1 = (s_3+c_3)^+ d_{12}$;
for Bob, $r_2 = (d_2+d_{12}+d_{23})^+d_{12}$; and
for Chris, $r_3 = (s_3+c_3)(s_3+c_3)(s_3+c_3)^+$.
These three requests are encoded using DFAs $\mathcal{D}_1$, $\mathcal{D}_2$, and $\mathcal{D}_3$, respectively. 

\changed{
The behavior of each guest is modeled by the event model in Figure~\ref{fig:party}a, in which $\mathbf{P}$ is the transition probability function of the model.
}
The joint behavior of the three guests is modeled by an event model $\mathcal{M}$ obtained as the Cartesian product of the models for the individuals, which has $6^3$ states in this example. 
The joint event model is further enhanced with joint events created from single events. For example, $d_{12}$ is the event in which Alice and Bob dance together.
To form a DFA $\mathcal{D}$ from the given specification DFAs, the robot uses
$\mathcal{D} = \mathbf{M_I}(\mathbf{M_I}(\mathbf{M_S}(\mathcal{D}_1),
\mathbf{M_S}(\mathcal{D}_2)), \mathbf{M_S}(\mathcal{D}_3))$.

Our implementation for this case study consists of 500 simulations for each of the settings \RTMFOM, \RTMFHM, and \RTM where the only observation is if Chris is currently smoking or not, which could perhaps be sensed through a smart smoke detector.
The expected number of steps for an optimal policy for \RTMFOM is 35.2, and over the 500 simulations, the average number of steps to record a story was 35.63, which is very close.
The average number for \RTM  with a single useful observation and \RTMFHM were respectively 37.68 and 40.4.
\vspace*{-6pt}
\section{Conclusions and future work}\label{sec:conc}
\vspace*{-8pt}
We have considered the problem of minimizing the expected time to record an event sequence satisfying a set of specifications. This was posed as the
problem of computing an optimal policy in an associated Markov decision problem.
Our implementation has verified that as the robot's ability to perceive the world increases, the expected number of steps to record a desired story decreases.
Future work should consider several extensions. For instance, factoring in the means needed to navigate in order to record an event, hence
the objective might minimize some expected cost rather than the expected number of steps. Or the case where a set of events
(rather than a single event), each assigned to a single robot, may be
predicted. Also the case where the robot is given new specification DFAs to
satisfy while recording stories for previous requests, especially where those specification DFAs are prioritized and the prioritization is subject to changes~\cite{rahmani2019optimal}, or perhaps the case
where the robot needs to learn a new event model to describe the environment owing 
to failure in predicting events.

\vspace*{-6pt}
\bibliography{manuscript}
  
\end{document}